\documentclass{article}

\newif\iffinal
\finaltrue
\iffinal
\usepackage[final]{nips_2016}
\else
\usepackage{nips_2016}
\fi

\usepackage[utf8]{inputenc} 
\usepackage[T1]{fontenc}    
\usepackage{hyperref}       
\usepackage{url}            
\usepackage{booktabs}       
\usepackage{amsfonts}       
\usepackage{nicefrac}       
\usepackage{microtype}      

\usepackage{graphicx} 
\usepackage{array}
\usepackage{amsmath}
\usepackage{placeins}
\usepackage{subcaption}
\usepackage{caption}
\usepackage{fancyhdr}
\usepackage{tabularx}

\setcitestyle{numbers,square}

\renewenvironment{abstract}%
{%
  \vskip 0.075in%
  \centerline%
  {\large\bf Abstract}%
  \vspace{0.5ex}%
  \begin{quote}%
}
{
  \par%
  \end{quote}%
  \vskip 1ex%
}

\date{}

\title{Recurrent Neural Networks With Limited Numerical Precision}

\author{
  Joachim Ott$^*$, Zhouhan Lin$^\ddagger$,Ying Zhang$^\ddagger$, Shih-Chii Liu$^*$, Yoshua Bengio$^\ddagger$$^\dag$ \\
  $^*$Institute of Neuroinformatics, University of Zurich and ETH Zurich \\
  \texttt{ottjoa+emdnn@gmail.com, shih@ini.ethz.ch} \\
  $^\ddagger$D\'epartement d'informatique et de recherche op\'erationnelle, Universit\'e de Montr\'eal \\
  $^\dag$CIFAR Senior Fellow \\
  \texttt{\{zhouhan.lin, ying.zhang\}@umontreal.ca} \\
}

\begin{document}
\maketitle

\begin{abstract}
Recurrent Neural Networks (RNNs) produce state-of-art performance on many machine learning tasks but 
their demand on resources in terms of memory and computational power are often
high. Therefore, there is a great interest in optimizing the computations performed
with these models especially when considering development of specialized
low-power hardware for deep networks. One way of reducing the computational needs is to limit the
numerical precision of the network weights and biases, and this will be addressed for the case of RNNs. We present results from the use of different stochastic and deterministic reduced precision training methods applied to two major RNN types, which are then tested
on three datasets. The results show that the stochastic and
deterministic ternarization, pow2-ternarization, and exponential quantization methods gave rise to low-precision RNNs that produce similar and even higher accuracy on certain datasets, therefore providing a path towards training more efficient implementations of RNNs
in specialized hardware.
\end{abstract}
\vspace*{-6mm}
\section{Introduction}
\vspace*{-3mm}A Recurrent Neural Network (RNN) is a specific type of neural network which is able to process
input and output sequences of variable length and therefore well suitable for sequence modeling. Various RNN architectures have been proposed in recent years based on different forms of non-linearity, such as the Gated Recurrent Unit (GRU) \cite{Cho2014} and Long-Short Term Memory (LSTM) \cite{Hochreiter1997}. They have enabled new levels of performance in many tasks including speech recognition \cite{Amodei2015}\cite{Chan2015} and machine translation \cite{Devlin2014}\cite{Chung2016}\cite{Sutskever2014}.

Compared to standard feed-forward networks, RNNs often take longer to train and are more demanding in memory and computational power. Thus, it is of great importance to accelerate computation and reduce training time of such networks. \\
Previous work showed the successful application of stochastic rounding strategies on feed forward networks, including binarization \cite{Courbariaux2015} and ternarization \cite{Lin2015} of weights of vanilla Deep Neural Networks (DNNs) and Convolutional Neural Networks (CNNs) \cite{Rastegari2016}, and in \cite{Courbariaux2016} even the quantization of their activations during training and run-time.  
Quantization of RNN weights has so far only been used with pretrained models \cite{shin2016fixed}.  

In this paper, a condensed and updated version of \cite{ott2016recurrent},	we use different methods to reduce the numerical precision of
weights in RNNs and test their performance on different benchmark
datasets. We use two popular RNN models: vanilla RNNs, and GRUs. Section
\ref{lowprecision} covers the three ways of obtaining low-precision weights
for the RNN models in this work, and Section \ref{exp} elaborates on the test results of the 
low-precision RNN models on different datasets. In \cite{ott2016recurrent} we additionally provide results and a possible explanation why binarization of RNNs does not work.  We make the code for the rounding methods available. \footnote{\url{https://github.com/ottj/QuantizedRNN}}
\section{Rounding Network Weights}  
\vspace*{-3mm}
\label{lowprecision}

This work evaluates the use of 3 different rounding methods on the weights of
two types of RNNs. These methods include the stochastic and 
deterministic ternarization method (TernaryConnect)
\cite{Lin2015}, the pow2-ternarization method \cite{Stromatias2015},
and exponential quantization \cite{ott2016recurrent}. For all 3 methods, we keep a full-precision copy of the
weights and biases during training to accumulate the small updates, while, during test time, we
can either use the learned full-precision weights or use their deterministic
low-precision version. As experimental results in Section \ref{exp} show, the
network with learned full-precision weights usually yields better results than
a baseline network with full precision during training, due to the extra
regularization effect brought by the rounding method. The deterministic
low-precision version could still yield comparable performance while
drastically reducing computation and required memory storage at test time. We
will briefly describe the former 3 low-precision methods, and elaborate on a additional method called Exponential Quantization we introduced in \cite{ott2016recurrent}.

\vspace*{-1mm}
\subsection{ Ternarization and Pow2-Ternarization}
\vspace*{-1mm}
TernaryConnect was first introduced in \cite{Lin2015}. By limiting the weights
to only 3 possible values, i.e., -1, 0 or 1
, this method does not require the use of multiplications. In
the stochastic version of the method, the low-precision weights are obtained
by stochastic sampling, while in the deterministic version, the weights are
obtained by thresholding.

TernaryConnect allows weights to be set to zero. Formally, the stochastic form
can be expressed as
\begin{equation}
W_{tern}=\text{sign}(W) \odot \text{binom}(p=\text{clip}(\text{abs}(2W),0,1))
\end{equation}
where $\odot $ is an element-wise multiplication, $\text{binom}(p)$ a function that draws samples from a binomial distribution, and $\text{clip}(x,a,b)$ clips values outside a given interval to the interval edges . In the deterministic form, the weights are quantized depending on 2 thresholds:
\begin{equation}
W_{tern}=  \left \{
  \begin{array}{ccc}
  +1 & \text{if} & W > 0.5 \\
  -1 &  \text{if} &  W \leq -0.5\\
    0 & \text{otherwise} &
  \end{array}
  \right.
\end{equation}

Pow2-ternarization is another fixed-point oriented rounding method
introduced in \cite{Stromatias2015}. The precision of fixed-point numbers is
described by the Q\textit{m.f}notation, where \textit{m} denotes the number of
integer bits including the sign bit, and \textit{f} the number of fractional
bits. For example, Q1.1 allows $(-0.5, 0, 0.5)$ as values.
The rounding procedure works as follows:
We first round the values to be in range allowed by the number of integer bits:
\begin{equation}
W_{\text{clip}} =  \left \{
  \begin{array}{ccc}
  -2^m & \text{where} & W \leq -2^m \\
  W & \text{where} & -2^m < W < 2^m \\
  2^m & \text{where} & W \geq 2^m 
  \end{array}
  \right.
\end{equation}
We subsequently round the fractional part of the values:
\begin{equation}
W_{p2t}=\text{round}(2^f \cdot W_{\text{clip}})\cdot 2^{-f}
\end{equation}

\vspace*{-1mm}
\subsection{Exponential Quantization}   \label{quantization}
\vspace*{-1mm}
Quantizing the weight values to an integer power of 2 is also a way of storing
weights in low precision and eliminating multiplications. Since quantization
does not require a hard clipping of weight values, it scales well with weight
values.

Similar to the methods introduced above, we also have a deterministic and
stochastic way of quantization. For the stochastic quantization, we sample the
logarithm of weight value to be its nearest 2 integers, with the probability of
getting one of them being proportional to the distance of the weight value from
that integer. For weights with negative weight values we take the logarithm of its absolute vale, but add their sign back after quantization. i.e.:
\begin{equation}
\label{quantize}
log_2{\left| W_b \right|}=  \left \{
  \begin{array}{ccc}
  \left\lceil log_{2}{\left| W \right|} \right\rceil & \text{with probability} & p=\frac{\left| W \right|}{2^{\left\lfloor log_{2}{\left| W \right|} \right\rfloor}}-1 \\
  \left\lfloor  log_{2}{\left| W \right|} \right\rfloor & \text{with probability} & 1-p
  \end{array}
  \right.
\end{equation}

For the deterministic version, we set $log_2{W_b}=\left\lceil log_{2}{W} \right\rceil$ if the $p$ in Eq. \ref{quantize} is larger than 0.5. 

Note that we just need to store the logarithm of quantized weight values. The
actual instruction needed for multiplying a quantized number differs according
to the numerical format. For fixed point representation, multiplying by a
quantized value is equivalent to binary shifts, while for floating point
representation it is equivalent to adding the quantized number's exponent to
the exponent of the floating point number. In either case, no complex operation
like multiplication is needed.

\vspace*{-2mm}

\vspace*{-2mm}
\section{Experimental Results and Discussion} \label{exp}
\vspace*{-3mm}

In the following experiments, we test the effectiveness of the different rounding methods on two different types of applications: character-level language modeling and speech recognition. 
\vspace*{-1mm}
\subsection{Vanilla RNN}
\vspace*{-1mm}
We validate the low-precision vanilla RNN on 2 datasets: text8 and Penn Treebank Corpus (PTB).

The \textbf{text8} dataset contains the first 100M characters from Wikipedia, excluding
all punctuations. It does not discriminate between cases, so its alphabet has
only 27 different characters: the 26 English characters and space. We take the
first 90M characters as training set and split them equally into sequences
with 50 character length each. The last 10M characters are split equally to
form validation and test sets. 

The \textbf{Penn Treebank Corpus} \cite{Taylor2003} (PTC) contains 50 different characters,
including English characters, numbers, and punctuations. We follow the settings
in \cite{mikolov2012subword} to split our dataset, i.e., 5017k characters for
training set, as well as 393k and 442k characters for validation and test set,
respectively.

\paragraph{Model and Training}
The models are built to predict the next character given the previous ones, and
performances are evaluated with the bits-per-character (BPC) metric, which is $\log_2$ of
the perplexity, or the per-character log-likelihood (base 2).
We use a RNN with ReLU activation and 2048 hidden units. 
We initialize hidden-to-hidden weights as identity matrices, while
input-to-hidden and hidden-to-output matrices are initialized with uniform noise.

\begin{figure}
          \vspace*{-3mm}
    \centering
    \begin{subfigure}[b]{0.48\textwidth} 
    \includegraphics[width=1\textwidth]{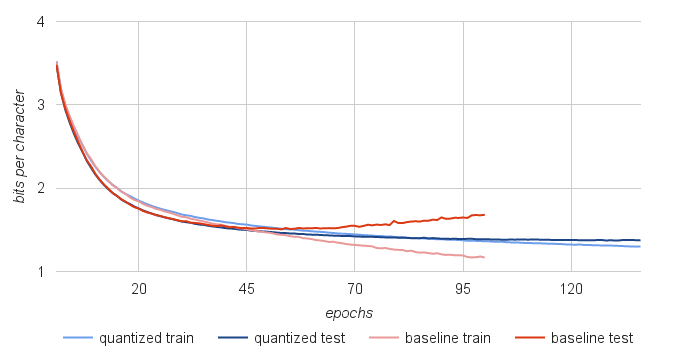}
    \label{fig:ptb}
    \caption{} \label{fig:ptbandtext8a}
    \end{subfigure}
    \begin{subfigure}[b]{0.48\textwidth} 
    \includegraphics[width=1\textwidth]{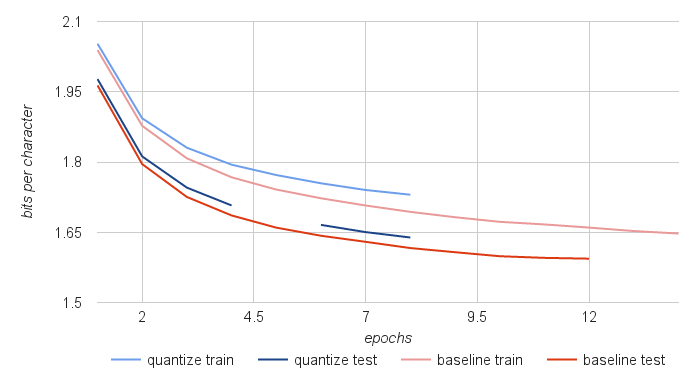}
	\label{fig:text8}
    \caption{} \label{fig:ptbandtext8b}
    \end{subfigure}
        \vspace*{-3mm}
    \caption{Training and test set performance of vanilla RNNs learning on PTB (a) and text8 (b) datasets.}
    \label{fig:ptbandtext8}
\end{figure}

We can see the regularization effect of stochastic quantization from the
results of the two datasets. 
In the PTB dataset, where the model size slightly overfits the dataset, the
low-precision model trained with stochastic quantization yields a test set
performance of 1.372 BPC that surpasses its full precision baseline (1.505
BPC) by around 0.133 BPC (Fig.~\ref{fig:ptbandtext8}, left). From the Figure \ref{fig:ptbandtext8} we can see
that stochastic quantization does not significantly hurt training speed and
manages to get better generalization when the baseline model begins to overfit.
On the other hand, we can also see, from the results on the text8 dataset where
the same sized model now underfits that the low-precision model performs worse
(1.639 BPC) than its baseline (1.588 BPC). (Fig. ~\ref{fig:ptbandtext8b} and
Table \ref{table:roundingResults}).

\begin{table}[ht]
\centering
\caption{Results from Vanilla RNNs (VRNNs) with exponential quantization.}
\label{table:roundingResults}
{\begin{tabular}{l l l l }
\hline
\textbf{Dataset} & \textbf{RNN Type} & \textbf{Baseline}  & \textbf{Exponential Quantization}  \\ \hline  \hline
text8 & VRNN & 1.588 BPC &  1.639 BPC   \\ \hline  \hline
PTC & VRNN & 1.505 BPC &  1.372 BPC  \\ \hline

\end{tabular}}

\end{table}

\vspace*{-1mm}
\subsection{GRU RNNs}
\vspace*{-1mm}
This section presents results from the various methods to limit numerical precision in the weights and biases of GRU RNNs which are then tested on the TIDIGITS dataset.

\paragraph{Dataset}
TIDIGITS\cite{Leonard1993} is a speech dataset consisting of clean speech of spoken numbers from 326 speakers.
We only use single digit samples (zero to nine) in our experiments, giving us 2464 training samples and 2486 validation samples. The labels for the spoken `zero' and `O' are combined into one label, hence we have 10 possible labels. We create MFCCs from the raw waveform and do leading zero padding to get samples of matrix size 39x200. The MFCC data is further whitened before use. 

\begin{figure}
          \vspace*{-3mm}
    \centering
    \begin{subfigure}[b]{0.49\textwidth}        
\includegraphics[trim = 20mm 80mm 20mm 100mm, clip,width=1\textwidth]{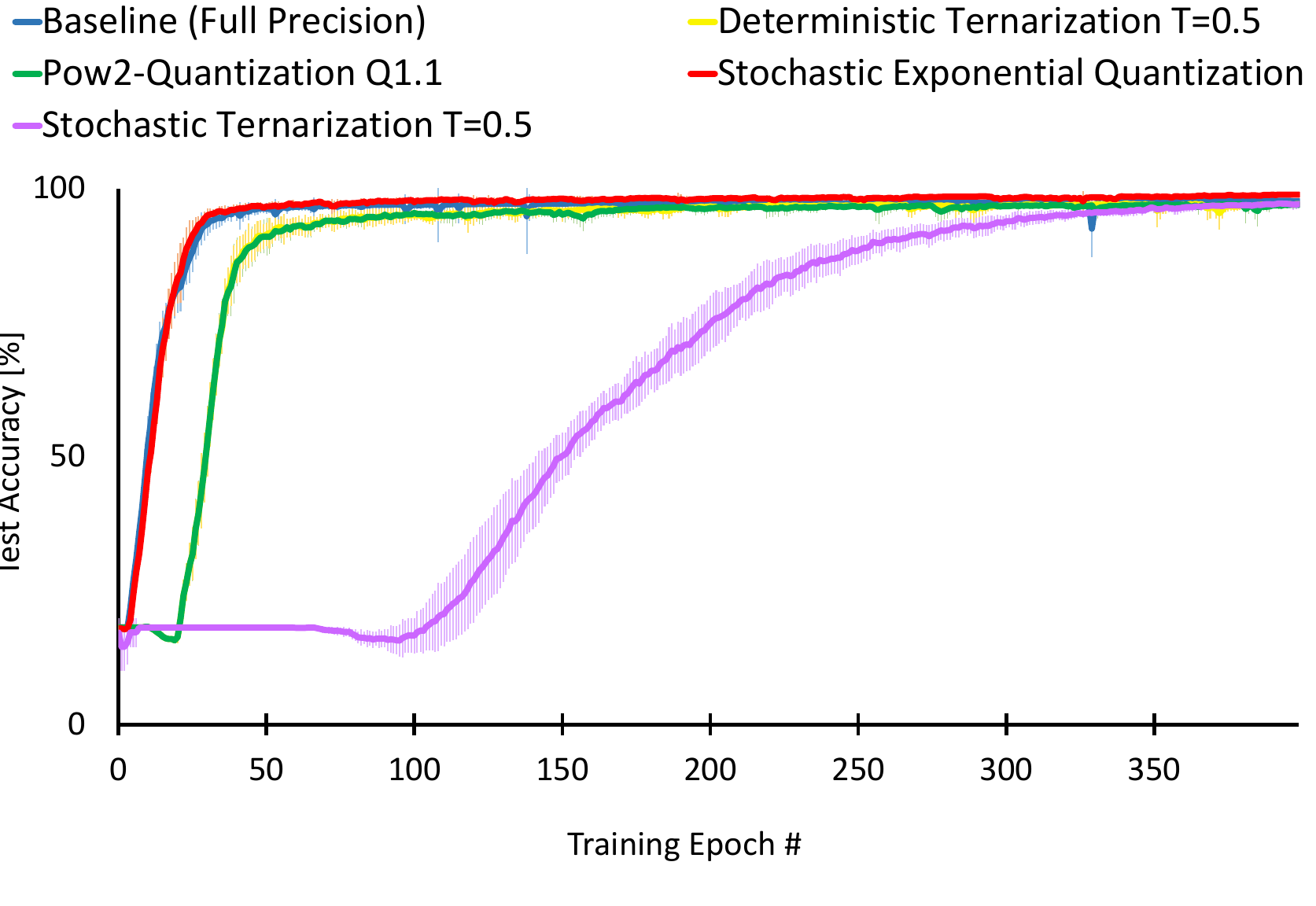}

\label{subfig:EffectDual-CopyRoundingConvergencePerformance_val_acc}
\caption{}

    \end{subfigure}
     ~
    \begin{subfigure}[b]{0.49\textwidth}
        \includegraphics[trim = 10mm 60mm 10mm 70mm, clip,width=1\textwidth]{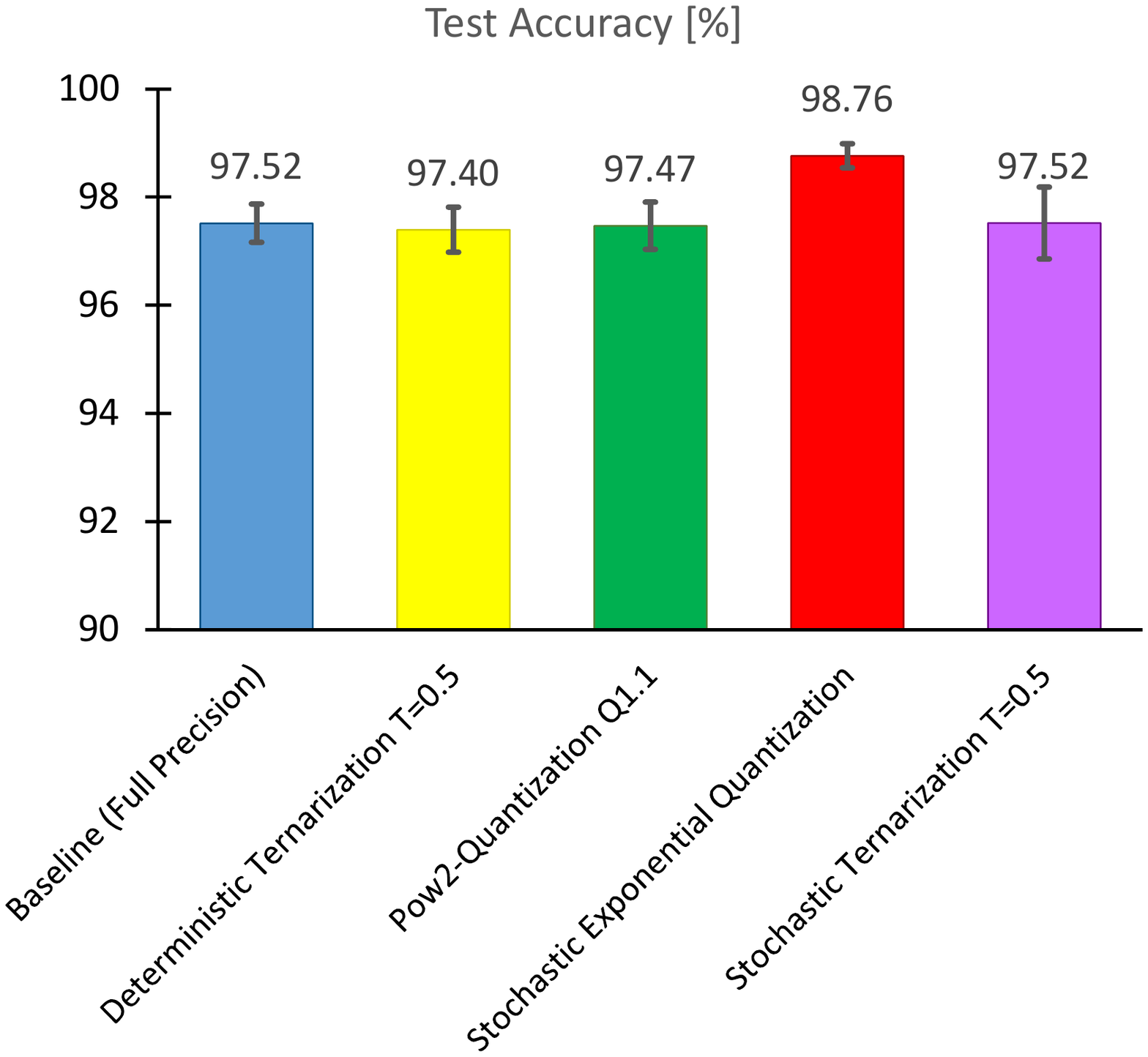}
     
        \label{fig:TernarizationGRUComponents_val_acc}
        \caption{}

    \end{subfigure}
           
    \caption{GRU models trained on the TIDIGITS speech recognition task. (a) Effect of rounding methods applied on GRU weights and biases during training compared against baseline (blue). Thick curves show the mean of 10 runs, half-transparent areas around a curve show the variance. 
(b) Mean final test accuracy of 10 runs per method. Exponential quantization applied on GRU weights and biases leads to higher than baseline final accuracy.
}
    \label{fig:GruComponents}
\end{figure}

\paragraph {Model and Training} The model has a 200 unit GRU layer followed by a 200 unit fully-connected ReLU layer. The output is a 10 unit softmax layer. Weights are initialized using the Glorot \& Bengio method \cite{Glorot2010}. The network is trained using Adam \cite{Kingma2014}, and BatchNorm \cite{Ioffe2015} is not used. 
We train our model for up to 400 epochs with a patience setting of 100 (no early stopping before epoch 100). GRU results in Figure \ref{fig:GruComponents} are from 10 experiments, with each experiment starting with a different random seed. This is done because previous experiments have shown that different random seeds can lead up to a few percent difference in final accuracy. We show average and maximum achieved performance on the validation set.

\paragraph{Effect of Rounding Methods}
To assess the impact of weight quantization, we trained our model and applied the rounding methods to all GRU weights and biases during training. Figure \ref{fig:GruComponents} (a) shows how rounding applied to GRU weights and biases has an effect on convergence compared to the full-precision baseline. Figure \ref{fig:GruComponents} (b) shows the mean final accuracy compared to baseline. These results show that limiting the numerical precision of GRU weights and biases using exponential quantization is beneficial for learning in this setup: the mean final accuracy is higher than baseline, while the convergence is as fast as baseline.

\vspace*{-2mm}
\section{Conclusion and Outlook}
\vspace*{-1mm}
This paper shows how low-precision quantization of weights can be performed effectively for RNNs.
We presented 3 existing methods of limiting the numerical precision, and used them on two major RNN types and determined how the limited numerical precision affects network performance across 3 datasets.
In the language modeling task, the low-precision model surpasses its full-precision baseline by a large gap (0.133 BPC) on the PTB dataset. We also show that the model will work better if put in a slightly overfitting setting, so that the regularization effect of stochastic quantization will begin to function. 
In the speech recognition task, we show that with stochastic ternarization we can achieve baseline results, and with exponential quantization even surpass the full-precision baseline.
The successful outcome of these experiments means that lower resource requirements are needed for custom implementations of RNN models.

\iffinal
\section{Acknowledgments}
We are grateful to INI members Danny Neil, Stefan Braun, and Enea Ceolini, and MILA members Philemon Brakel, Mohammad Pezeshki, and Matthieu Courbariaux, for useful discussions and help with data preparation. 
We thank the developers of Theano\cite{2016arXiv160502688short}, Lasagne, Keras, Blocks\cite{VanMerrienboer2015}, and Kaldi\cite{Povey_ASRU2011}.\\
The authors acknowledge partial funding from the Samsung Advanced Institute of Technology, University of Zurich, NSERC,
CIFAR and Canada Research Chairs.\\
\fi

\small
\FloatBarrier

\bibliographystyle{plainnat}
\bibliography{ott_lin_biblio}
\end{document}